%% file: erk.tex
\documentclass[a4paper]{article}
\usepackage{cite}

\usepackage[utf8]{inputenc}
\usepackage{erk}
\usepackage{times}
\usepackage{graphicx}
\usepackage[top=22.5mm, bottom=22.5mm, left=22.5mm, right=22.5mm]{geometry}

\usepackage{amsmath}
\usepackage{amssymb}
\usepackage{amsthm}
\usepackage{amsfonts}  
\usepackage{subcaption}
\usepackage{caption}
\usepackage{booktabs} 
\usepackage{tabularx}
\usepackage{pifont}
\usepackage{xcolor}
\usepackage{multirow}  
\usepackage{array}     

\newcommand{\green}[1]{\textcolor{green}{#1}}
\newcommand{\red}[1]{\textcolor{red}{#1}}
\newcommand{\R}{\mathbb{R}}

\newcommand{\D}{d}

\newcommand{\tensor}[1]{\boldsymbol{#1}}
\newcommand{\subfigwidth}{0.25}

\usepackage[slovene,english]{babel}

\def\footnotemark{}

\begin{document}
\title{Mere prekrivanja večrazsežnih Gaussovih rojev v nenadzorovanem sprotnem učenju}

\thanks{Prvi avtor se zahvaljuje Javni agenciji za znanstvenoraziskovalno in inovacijsko dejavnost Republike Slovenije (Slovenian Research and Innovation Agency -- ARIS) za finančno podporo, projekt P2-0219.}%

\author{Miha Ožbot$^{1}$, Igor Škrjanc$^{2}$} 

\affiliation{$^{1,2}$Fakulteta za elektrotehniko, Univerza v Ljubljani, Slovenija}
\email{$^{1}$miha.ozbot@fe.uni-lj.si, $^{2}$igor.skrjanc@fe.uni-lj.si}
\maketitle
\begin{abstract}{Measures of Overlapping Multivariate Gaussian Clusters in Unsupervised Online Learning}
In this paper, we propose a new measure for detecting overlap in multivariate Gaussian clusters. The aim of online learning from data streams is to create clustering, classification, or regression models that can adapt over time based on the conceptual drift of streaming data. In the case of clustering, this can result in a large number of clusters that may overlap and should be merged. Commonly used distribution dissimilarity measures are not adequate for determining overlapping clusters in the context of online learning from streaming data due to their inability to account for all shapes of clusters and their high computational demands. Our proposed dissimilarity measure is specifically designed to detect overlap rather than dissimilarity and can be computed faster compared to existing measures. Our method is several times faster than compared methods and is capable of detecting overlapping clusters while avoiding the merging of orthogonal clusters.
\end{abstract}

\selectlanguage{slovene}

\section{Uvod}

\par Samorazvijajoči se modeli \cite{Skrjanc_Iglesias_Sanchis_Leite_Lughofer_Gomide_2019} so matematični modeli, katerih naloga je sprotno prilagajanje strukture in parametrov za namen napovedovanja odvisne spremenljivke, rojenja podatkov ali klasifikacije v razrede. Opazovan sistem se nahaja v spremenljivem okolju, zato je podvržen konceptualnemu lezenju in skočnim spremembam parametrov. Takšne sisteme pogosto srečujemo v realnem svetu, od spremenljivih razmer v industrijskih obratih, finančnih transakcijah in naložbah, do medicinskih podatkovnih tokov, socialnih omrežij in kibernetske varnosti. Ključna zahteva je, da se model lahko hitro prilagodi spremembam in je računsko učinkovit, saj je količina podatkov v primeru podatkovnih tokov neomejena. Ti sistemi običajno uporabljajo metode nenadzorovanega rojenja podatkov za določitev strukture in izluščenje informacij iz podatkov. V ta namen se pogosto uporabljajo Gaussovi roji~\cite{Lughofer_Škrjanc_2023, Ozbot_Lughofer_skrjanc_2023, Skrjanc_2020, Dovzan_Logar_Škrjanc_2015, Pratama_Anavatti_Joo_Lughofer_2015}, ki so predstavljeni kot multivariantne normalne distribucije (\textit{multivariate normal distributions}) s kovariančnimi matrikami podatkov $P \sim \mathcal{N}(\tensor{\mu}, \tensor{\Sigma})$. Z večjim številom rojev želimo tako aproksimirati tudi bolj zapletene distribucije podatkov, čeprav opazovani proces ne sledi Gaussovi porazdelitvi.

\par V primeru sprotnega učenja iz toka podatkov nastane večje število rojev, ki se začnejo prekrivati s povečevanjem količine podatkov. Občasno je potrebno te roje združiti, da zmanjšamo računsko zahtevnost metode in poenostavimo model. Pri tem se pojavlja ključno vprašanje: kako najti prekrivajoče se roje med sprotnim učenjem čim bolj učinkovito. V literaturi je bilo predlaganih več metod za detekcijo prekrivanja rojev, vendar nobena od teh metod še ni rešila težave računske učinkovitosti in detekcije prekrivanja v vseh primerih \cite{Pratama_Anavatti_Joo_Lughofer_2015, Dovzan_Logar_Škrjanc_2015, Skrjanc_2020}. Težava teh mer je, da so osredotočene na podobnost distribucij in ne na prekrivanje rojev, zato ne uspejo zaznati prekrivanja rojev različnih velikosti. Potencialno bi lahko izboljšali hitrost združevanja z uporabo paralelnega procesiranja, kar bi omogočilo združevanje večjega števila rojev v enem koraku, vendar to še ni bilo raziskano v literaturi. Drugo ključno vprašanje je, ali je bolj učinkovito združevati roje postopoma po dva ali več rojev hkrati. V prvem primeru je potrebno večkrat preverjati prekrivanje rojev, medtem ko je v drugem primeru treba najti najboljšo kombinacijo rojev za združevanje. Zaželeno bi bilo združiti čim večje število rojev v enem koraku, saj je to najučinkovitejši pristop za implementacijo izračuna na grafičnih karticah. Težava je, da je iskanje največjih skupin (\textit{clique}) rojev za združevanje po svoji naravi rekurziven problem in še ne obstaja algoritem, ki bi našel optimalno rešitev v polinomskem času (\textit{NP-hard}). Pri tem ima potencialna uporaba grafičnih kartic \cite{Almasri_Chang_Hajj_Nagi_Xiong_Hwu_2023, Wei_Chen_Tsai_2021}, zaradi paralelnega procesiranja potencial pospešiti iskanje skupnosti v velikih grafih. Klasične metode so temeljile na rekurziji, ki pa ni primerna za računanje na grafični kartici.
\par Cilj študije je razviti metodo za detekcijo prekrivajočih se rojev, ki omogoča stabilen in hiter izračun za visoke dimenzije in veliko število rojev. Zanima nas, ali je mogoče procesiranje prekrivanja in združevanja rojev izvesti vzporedno, kar bi omogočilo učinkovito uporabo grafičnih kartic. Te so se v zadnjem desetletju izkazale kot bolj učinkovite od izračunov na procesorjih, zlasti če lahko izvedemo veliko število matričnih operacij brez rekurzivnih zank ter brez shranjevanja ali sinhronizacije vmesnih korakov. V tem članku predstavljamo novo mero prekrivanja rojev, ki temelji na razmerju volumnov rojev. Ta metoda je računsko učinkovita in stabilna pri višjih dimenzijah, omogoča upoštevanje števila vzorcev v rojih, zaznava majhne roje in izloča roje različnih oblik.

\section{Metodologija}
\par Za združevanje rojev potrebujemo mero prekrivanja rojev, postopek izbire rojev, ki so primerni za združevanje, ter izračun kovariančne matrike združenega roja. Roj, ki ga indeksiramo z $i \,{\in}\, \mathcal{I}$, pri čemer je $\mathcal{I} \,{=}\, \{1, \ldots, c\}$, je popolnoma definiran s številom vzorcev $n_i$, ki mu pripadajo, kovariančno matriko $\tensor{\Sigma}_i$ in pričakovano vrednostjo ali središčem roja $\tensor{\mu}_i$. Zanima nas mera, ki najbolje opisuje prekrivanje rojev in je hkrati računsko najbolj učinkovita.
\subsection{Mere podobnosti distribuciji}
\par Naj bosta $P \sim \mathcal{N}(\tensor{\mu}_P, \tensor{\Sigma}_P)$ in $Q \sim \mathcal{N}(\tensor{\mu}_Q, \tensor{\Sigma}_Q)$ multivariantni Gaussovi porazdelitvi, ki definirata primerjana roja. Obravnavamo mere podobnosti in neskladnosti (\textit{dissimilarity}), ki služijo kot mere prekrivanja rojev.

\subsubsection{Razdalja Bhattacharyya}
\par Razdalja Bathacharyya (B) je pogosto uporabljena kot mera prekrivanja rojev, ker je simetrična in posredno normira razdalje glede na povprečno kovariančno matriko rojev~\cite{Lughofer_Cernuda_Kindermann_Pratama_2015, Lughofer_Pratama_Skrjanc_2018, Baidari_Honnikoll_2021}: 
\begin{align}
    D_B(P{\parallel}Q) =  & \frac{1}{8} (\tensor{\mu}_Q {-} \tensor{\mu}_P)^\top \tensor{\Sigma_{M}}^{-1} (\tensor{\mu}_Q {-} \tensor{\mu}_P) \notag + \\  
    & \frac{1}{2} \ln \left( \frac{\det \tensor{\Sigma}_M}{\sqrt{\det \tensor{\Sigma}_P \det \tensor{\Sigma}_Q}} \right),
\end{align}
kjer je $\tensor{\Sigma}_M = \frac{1}{2}(\tensor{\Sigma}_P {+} \tensor{\Sigma}_Q)$.
\par Namesto determinant lahko razpišemo izraz v vsoto logaritmov determinant, katerih izračuni so veliko bolj numerično stabilni. Zanimiva lastnost drugega člena je, da je števec enak determinanti povprečja in imenovalec geometrično povprečje determinant kovariančnih matrik. 
\subsubsection{Divergenca Jensen-Shannon}
\par Divergenca Jensen-Shannon (JS) je razširitev divergence Kullback-Leibler (KL), ki rešuje problem nesimetričnosti divergence KL z uvedbo združene distribucije \( M \sim \mathcal{N}(\tensor{\mu}_M, \tensor{\Sigma}_M) \), ki predstavlja povprečje obeh distribucij in omogoča simetrično merjenje razdalje med njima~\cite{Pardo_2005}:
\begin{align}
D_{JS}(P {\parallel} Q)  &= \frac{1}{2} \big(D_{KL}(P {\parallel} M) + D_{KL}(Q {\parallel} M)\big), \\ 
D_{KL} (P{\parallel} M)  & = \frac{1}{2}\Bigg(\operatorname{tr}\left(\tensor{\Sigma}_M^{-1} \tensor{\Sigma}_P\right) + \ln \left(\frac{\operatorname{det} \tensor{\Sigma}_M}{\operatorname{det}\tensor{\Sigma}_P}\right) + \notag \\ &  \left(\tensor{\mu}_M{-}\tensor{\mu}_P\right)^\top \tensor{\Sigma}_M^{-1}\left(\tensor{\mu}_M{-}\tensor{\mu}_P \right) - \D\Bigg) ,
\end{align}
kjer je $\tensor{\mu}_M = \frac{1}{2}(\tensor{\mu}_P {+} \tensor{\mu}_Q)$ in $\tensor{\Sigma}_M = \frac{1}{2}(\tensor{\Sigma}_P {+} \tensor{\Sigma}_Q)$.
\subsubsection{Razdalja (2-)Wassersteina}
\par Razdalja Wassersteina drugega reda (W), ki temelji na kvadratu razdalje med porazdelitvama, je mera stroška optimalnega prenosa mase iz ene porazdelitve v drugo~\cite{Salmona_Delon_Desolneux_2021}:
\begin{equation}
\begin{split}
D_W&(P \parallel Q) = \left\|\tensor{\mu}_P - \tensor{\mu}_Q\right\|^2 + \\  
& \operatorname{Tr}\left(\tensor{\Sigma}_P + \tensor{\Sigma}_Q - 2\left(\tensor{\Sigma}_P^{1/2} \tensor{\Sigma}_Q \tensor{\Sigma}_P^{1/2}\right)^{1/2}\right),
\end{split} 
\end{equation}

\subsubsection{Mera prekrivanja eGauss}
\par Mera prekrivanja rojev eGauss+~\cite{Skrjanc_2020} temelji na razmerju volumnov pred in po združitvi rojev:
\begin{equation}
D_{eGauss+}(P{\parallel} Q) = \frac{\det(\tensor{\Sigma}_{M})}{\det(\tensor{\Sigma}_P) + \det(\tensor{\Sigma}_Q)}.
\end{equation} 
Pri tem je združena kovariančna matrika je enaka:
\begin{align}
n_{M} & = n_P + n_Q,\label{eq_merging_increment_samples}\\
    \tensor{\mu}_{M} & = \big({n_{P} \tensor{\mu}_P + {n}_Q \tensor{\mu}_Q}\big)/{n_{M}}, \label{eq_merging_centers} \\
      \tensor{\Sigma}_{M}({n_{M}{-}1}) &= 
     (n_P{-}1)\tensor{\Sigma}_P + (n_Q{-}1)\tensor{\Sigma}_Q +  \notag \\ & \tensor{M}^{\top}_{P}\tensor{E}^{\top}_{P}\tensor{E}_{P}\tensor{M}_{P} + \tensor{M}^{\top}_{Q}\tensor{E}^{\top}_{Q}\tensor{E}_{Q}\tensor{M}_{Q} - \notag\\ &\tensor{M}^{\top}_{M}\tensor{E}^{\top}_{M}\tensor{E}_{M}\tensor{M}_{M}, \label{eq_egauss_zdruzevanje} 
\end{align}
kjer $\tensor{M}_P {=} \tensor{\mu}^{\top}_P\mathrm{I}\,{\in}\,\R^{\D\times \D}$ predstavlja diagonalno matriko, ki na diagonali vsebuje spremenljivke središč rojev, $\D$ je dimenzija problema ali število značilk ter $\tensor{E}_{P}\,{\in}\,\R^{n_P\times \D}$ je matrika, v kateri so vsi elementi enaki 1, enako velja za $Q$ in $M$.
\par Ta mera omogoča izključitev rojev, ki imajo ortogonalne lastne vektorje, vendar ne vključuje zaščite za visoke dimenzije kovariančnih matrik. Ortogonalnost v tem kontekstu pomeni, da so lastne vrednosti, ki definirajo roje, med seboj razmeroma pravokotne v prostoru značilk. Zanimiv je tudi postopek združevanja rojev, ki ni več preprosto povprečje rojev, temveč upošteva definicijo kovariančne matrike in število vzorcev v roju. To omogoča, da se večjemu številu vzorcev v roju pripisuje večji vpliv, kar rešuje problem, pri katerem bi distribucija ali roj z majhnim številom vzorcev prevladal pri izračunu, čeprav ima druga mera veliko večje število vzorcev.

\subsubsection{Predlagana mera prekrivanja}
\par Želimo mero, ki meri prekrivanje rojev (ne podobnosti) in omogoča detekcijo različnih velikosti rojev (I), je računsko stabilna pri visokih dimenzijah $\D$ (II), in je računsko učinkovita (III). Prvo zahtevo smo rešili z uporabo aritmetičnega povprečja determinante rojev, kot pri meri eGauss+. Drugo zahtevo lahko rešimo tako, da namesto determinante $\det(\Sigma)$ izračunamo $\ln(\det(\Sigma))$. Tako dobimo mero prekrivanja (\textit{overlapping}):
\begin{align} 
   &D_{O} (P{\parallel} Q) = \ln \left( \frac{\det(\tensor{\Sigma}_M)}{\frac{1}{2}\left(\det(\Sigma_P) +  \det(\Sigma_Q)\right)} \right) 
 = \notag \\& \ln(2) {+} \ln\det(\tensor{\Sigma}_M) {-} \ln\!{\big(e^{\ln\det(\tensor{\Sigma}_P)} {+} e^{\ln\det(\tensor{\Sigma}_Q)}\big)},
\end{align}
kjer je združena kovariančna matrika definirana kot:
\begin{align}
{(n_p {+} n_q{-}1)}\tensor{\Sigma}_{M}  & = (n_{p} {-} 1)\tensor{\Sigma}_p + (n_{q} {-} 1)\tensor{\Sigma}_q + \notag\\ &  + \frac{n_{p}n_{q}}{n_p {+} n_q} (\tensor{\mu}_{p} {-} \tensor{\mu}_{q})(\tensor{\mu}_{p} {-} \tensor{\mu}_{q})^\top.
\end{align}
\par Enačba za združeno kovariančno matriko temelji na poenostavljenem zapisu enačbe (\ref{eq_egauss_zdruzevanje}). Za primerjavo smo preizkusili uporabo aritmetičnega povprečja še za mero Bathacharyya (aB). Mera Bhattacharyya se izkaže za zelo potencialno, vendar povzroča težave geometrično povprečje v imenovalcu drugega člena, če sta velikosti rojev zelo različni. To samo po sebi ni slabost, če uporabljamo razdaljo kot mero podobnosti, saj sta roja res različna, vendar ne omogoča detekcije prekrivajočih se rojev. To slabost lahko izničimo, če v meri zamenjamo geometrijsko povprečje z aritmetičnim povprečjem. Poleg tega se lahko izognemo izračunu inverza v prvem členu tako, da za novo združeno matriko uporabimo metodo združevanja rojev, ki vključuje informacijo o oddaljenosti središč rojev. 
Vrednost mere prekrivanja rojev je manjša od nič, ko je volumen združenega roja manjši od povrečja volumnov obeh rojev. 
\par Tretja zahteva predstavlja največji izziv, saj izračuni volumnov, lastnih vektorjev in inverzov matrik temeljijo na izračunu determinante, ki je računsko zahtevna operacija. To zahtevo rešujemo z oceno zgornjih mej za determinante namesto z izračunom točne vrednosti. Hadamardova neenakost pravi, da je za vsako pozitivno semi-definitno matriko $ \Sigma \in \mathbb{R}^{\D \times \D} $ determinanta matrike manjša ali enaka produktu njenih diagonalnih elementov~\cite{Różański_Wituła_Hetmaniok_2017}:
\begin{equation}
    \det(\Sigma) \leq \prod_{l=1}^{\D} \sigma_{ll}.
\end{equation}
Mera prekrivanja, ki uporablja Hadamardovo neenakost ($\tilde{\mathrm{O}}$), predstavlja oceno zgornje meje volumna rojev. V meri zamenjamo točni izračun determinante z oceno, kar drastično pohitri izračune za visoke dimenzije. Slabost tega pristopa je, da pri tem izgubimo sposobnost detekcije rojev z zelo različnimi lastnimi vrednostmi, saj izgubimo informacije o korelacijah med značilkami.

\subsection{Združevanje večjega števila rojev}
\par Učinkovitost združevanja rojev lahko izboljšamo tudi tako, da združimo večje število rojev v enem koraku, namesto da to izvajamo iterativno. Pri iterativnem združevanju je potrebno večkrat zaporedoma izračunati mere podobnosti, kar je lahko zelo neučinkovito, zlasti pri uporabi grafičnih kartic. Preverjanje prekrivanja med pari rojev je zahteven postopek, še posebej če je rojev veliko, vendar lahko uporabimo dodatne kriterije za izbiro ustreznih kandidatov. Preverjanje vseh možnih kombinacij rojev za različno število prekrivajočih se rojev je neučinkovito, če ne celo nemogoče. Predlagamo primerjavo parov rojev in iskanja največje skupine medsebojno prekrivajočih se rojev. Ključno pri tem je najti največje skupine (\textit{maximal cliques}) rojev, ki so primerne za združevanje. V tem članku ni poudarek na učinkovitosti izračuna največjih skupin, ampak na prikazovanju delovanja mere podobnosti in metode združevanja več rojev v enem računskem koraku. Uporabljamo matrični zapis sosednjih rojev (\textit{adjacency matrix}), ki smo ga dobili z našo mero prekrivanja, in poljuben klasičen algoritem za iskanje največje skupine.
\par Za združevanje večjega števila rojev potrebujemo novo enačbo. Predlagamo izboljšavo enačbe (\ref{eq_egauss_zdruzevanje}) za združene kovariančne matrike, ki omogoča združevanje več rojev. Naj bo $\mathcal{J} \,{\subseteq}\, \mathcal{I}$ množica rojev, ki jih želimo združiti. Izpeljava enačbe za združitev temelji na $\tensor{X}^\top_{M} \tensor{X}_{M} = \sum_{j\in\mathcal{J}}\tensor{X}^\top_{j} \tensor{X}_{j}$ in $\tensor{\Sigma}_{M} = \frac{1}{(n_{M} {-} 1)}(\tensor{X}^\top_{M} \tensor{X}_{M} {-} n_{M}\tensor{\mu}_{M}\tensor{\mu}_{M}^\top)$, kjer $\tensor{X}^\top_{M} = \big[\tensor{X}^\top_{j} : j \,{\in}\, \mathcal{J}\big] \in \R^{(\sum_jn_j)\times \D}$ predstavlja matriko podatkov, ki vsebuje vse vzorce rojev. To lahko formuliramo v enačbo, ki ne potrebuje matrik podatkov:
\begin{align}\label{eq_zdruzevanje}
&(n_{M} {-} 1)\tensor{\Sigma}_{M} = \notag \\ & \sum_{j \in \mathcal{J}} (n_j {-} 1) \tensor{\Sigma}_j 
+ \sum_{j \in \mathcal{J}} n_j \tensor{\mu}_j \tensor{\mu}_j^\top - n_{M}\tensor{\mu}_{M}\tensor{\mu}_{M}^\top = \\
&\underbrace{\sum_{j \in \mathcal{J}} (n_j - 1) \tensor{\Sigma}_j}_{\text{varianca}} + \underbrace{\sum_{i \in \mathcal{J}} \sum_{j \in \mathcal{J}, j > i} 
\frac{n_i n_j}{n_{M}} (\tensor{\mu}_i {-} \tensor{\mu}_j)(\tensor{\mu}_i {-} \tensor{\mu}_j)^\top}_{\text{razdalja med središči}}, \notag
\end{align}
kjer je \( n_{M} = \sum_{j \in \mathcal{J}} n_j \) in \( \tensor{\mu}_{M} = \frac{1}{n_{M}}\sum_{j \in \mathcal{J}}n_j\tensor{\mu}_{j} \).
\par Izpeljavo izpustimo zaradi obsežnosti. Končni zapis enačbe je bolj ilustrativen, vendar računsko manj učinkovit za večje število rojev kot začetni. Združen roj je sestavljen iz prispevkov kovariančnih matrik posameznih rojev ter razdalje med središči, ki je člen ranga ena. Ta zapis omogoča vizualizacijo vpliva posameznega člena na združeni roj.
\section{Eksperimentiranje}
\par
\begin{figure}[!htb]
    \centering
    \begin{subfigure}[b]{\subfigwidth\textwidth}
        \centering
        \includegraphics[width=1\linewidth]{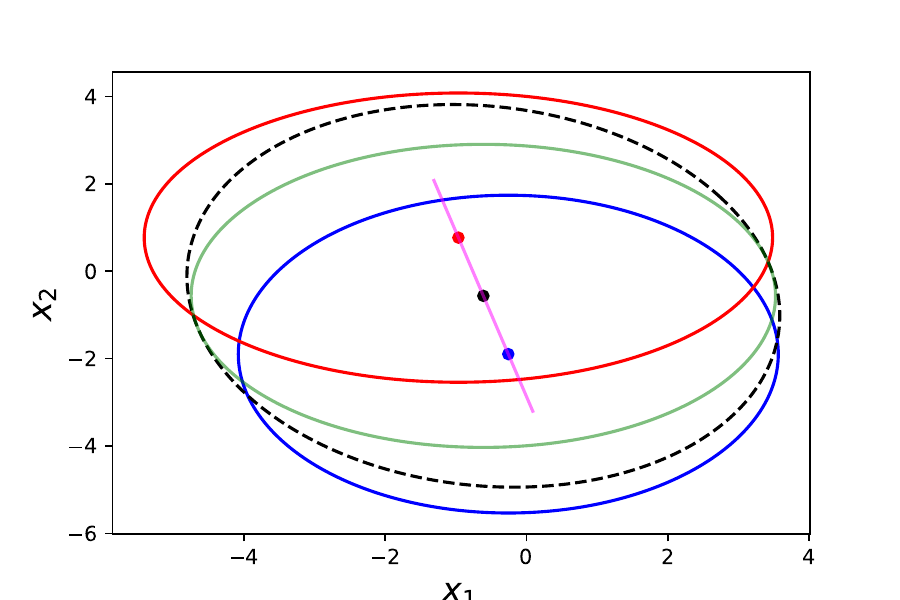}
        \caption{Prekrivanje rojev.} \label{fig_basic}
    \end{subfigure}
    \hspace{-0.03\textwidth} 
    \begin{subfigure}[b]{\subfigwidth\textwidth}
        \centering
        \includegraphics[width=1\linewidth]{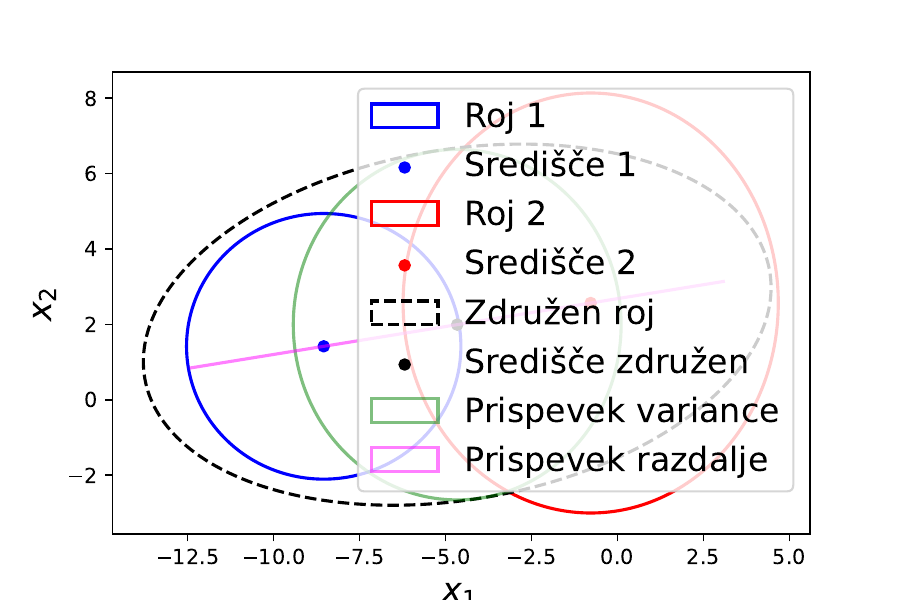}
        \caption{Ni prekrivanja rojev.} \label{fig_no_merge}
    \end{subfigure}
    

    \begin{subfigure}[b]{\subfigwidth\textwidth}
        \centering
        \includegraphics[width=1\linewidth]{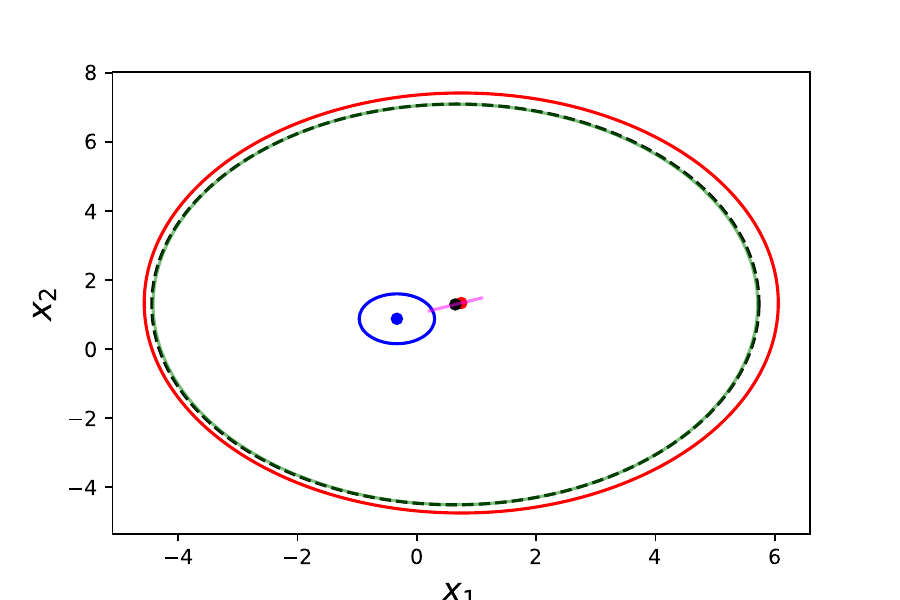}
        \caption{Velik in majhen roj.} \label{fig_small}
    \end{subfigure}
    \hspace{-0.03\textwidth} 
    \begin{subfigure}[b]{\subfigwidth\textwidth}
        \centering
        \includegraphics[width=1\linewidth]{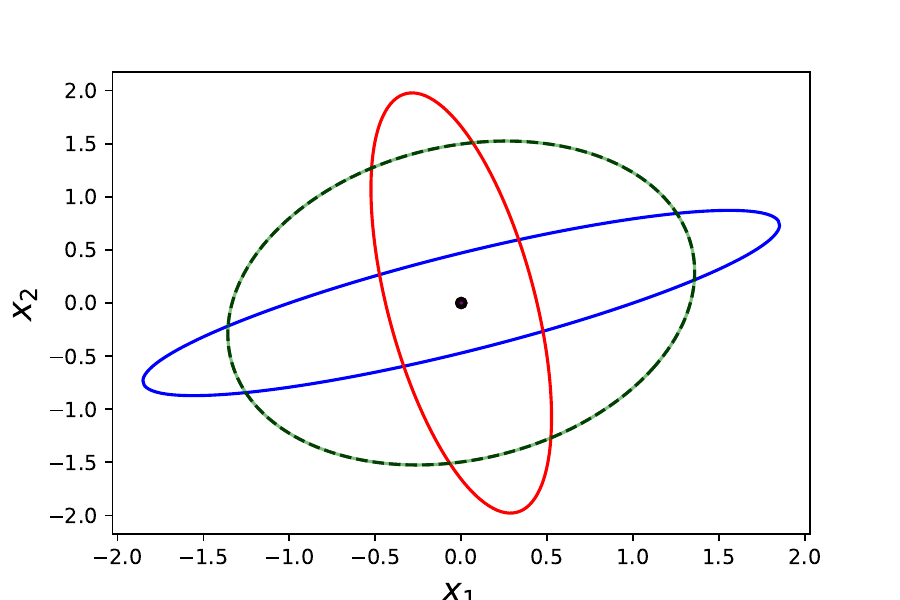}
        \caption{Pravokotna roja.} \label{fig_cross}
    \end{subfigure}
    \caption{Značilni primeri prekrivanja rojev pri sprotnem učenju, ki jih uporabljamo pri eksperimentiranju. Prikazana sta središči in 2$\sigma$ elipsi rojev pred združitvijo ter združeni roj. Prikazani so tudi ločeni prispevki zaradi razdalje med središči in prispevki kovariančnih matrik.}
     \label{slika_eskperimenti}
\end{figure}

 Naš interes je mera divergence, ki je zmožna zaznati vse primere prekrivanja, pogosto prisotne v problematiki sprotnega učenja v spremenljivem okolju. Ker je pri sprotnem učenju ključnega pomena hitrost metode zaradi velike količine podatkov v tokovih podatkov in zahtevnosti izračuna, nas zanima ne le točnost vrednosti divergence, temveč tudi računska zahtevnost posamezne mere, ki jo ocenjujemo s časom, potrebnim za izračun. Na sliki \ref{slika_eskperimenti} so predstavljeni različni scenariji združevanja rojev: (a) prekrivajoča se roja, (b) neprekrivajoča se roja, (c) majhen roj znotraj veliko večjega roja in (d) roja z zelo različnimi dominantnimi lastnimi vrednostmi, ki imata skupno središče. V primeru majhnega roja je zaželeno, da mera pokaže prekrivanje, medtem ko v primeru pravokotnih rojev želimo, da mera pokaže odsotnost prekrivanja. Roji so bili ustvarjeni naključno.

\begin{table*}[h!]
\centering
\scriptsize
\setlength{\tabcolsep}{2pt}
\begin{tabular}{@{}ll|cc|cc|cc|cc|cc|cc|cc@{}}
\toprule
$\D$ & \textbf{Experiment} & \multicolumn{2}{c}{\textbf{B}}  & \multicolumn{2}{c}{\textbf{JS}}  & \multicolumn{2}{c}{\textbf{W}}  & \multicolumn{2}{c}{\textbf{eGauss+}}  & \multicolumn{2}{c}{\textbf{aB(naša)}}  & \multicolumn{2}{c}{\textbf{O(naša)}}  & \multicolumn{2}{c}{$\mathbf{\tilde{\mathrm{O}}}$\textbf{(naša)}} \\& & \textbf{V} & \textbf{t [$\mu$s]} & \textbf{V} & \textbf{t [$\mu$s]} & \textbf{V} & \textbf{t [$\mu$s]} & \textbf{V} & \textbf{t [$\mu$s]} & \textbf{V} & \textbf{t [$\mu$s]} & \textbf{V} & \textbf{t [$\mu$s]} & \textbf{V} & \textbf{t [$\mu$s]} \\
\midrule
\multirow{4}{*}{2} & prekrivanje \ding{51} & \green{0.06} & 149(2.8x) & \green{0.06} & 79(1.5x) & \green{1.25} & 162571(3010.6x) & \green{0.48} & 217(4.0x) & \green{0.03} & 68(1.3x) & \green{0.04} & \textbf{54} & \green{0.05} & 55(1.0x) \\ 
 & ni prekrivanja \ding{55} & \green{1.89} & 87(1.7x) & \green{1.89} & 89(1.8x) & \green{10.92} & 296(5.9x) & \green{5.22} & 132(2.6x) & \green{1.86} & 65(1.3x) & \green{1.52} & \textbf{50} & \green{1.59} & 57(1.1x) \\ 
 & majhen roj \ding{51} & \red{1.50} & 110(1.4x) & \red{1.50} & 126(1.6x) & \red{13.83} & 386(4.9x) & \green{0.87} & 172(2.2x) & \green{-0.29} & \textbf{78} & \green{0.51} & 96(1.2x) & \green{0.51} & 92(1.2x) \\ 
 & pravokotna \ding{55} & \red{0.85} & 114(1.5x) & \red{0.85} & 120(1.6x) & \red{1.22} & 406(5.5x) & \green{2.72} & 152(2.1x) & \green{0.85} & \textbf{74} & \green{1.69} & 96(1.3x) & \green{0.84} & 86(1.2x) \\ 
\midrule
\multirow{4}{*}{100} & prekrivanje \ding{51} & \green{1.45} & 1006(5.7x) & \green{1.45} & 930(5.2x) & \green{14.95} & 2254(12.7x) & \green{23.63} & 1066(6.0x) & \green{1.19} & 567(3.2x) & \green{1.46} & 432(2.4x) & \green{1.88} & \textbf{178} \\ 
 & ni prekrivanja \ding{55} & \green{117.08} & 720(3.0x) & \green{117.08} & 922(3.8x) & \green{96.23} & 2942(12.3x) & \green{146.08} & 1687(7.0x) & \green{116.24} & 639(2.7x) & \green{4.27} & 588(2.5x) & \green{91.78} & \textbf{240} \\ 
 & majhen roj \ding{51} & \red{75.38} & 582(2.8x) & \red{75.38} & 838(4.0x) & \red{666.22} & 3115(14.8x) & \green{0.04} & 783(3.7x) & \green{-25.99} & 553(2.6x) & \green{-7.84} & 467(2.2x) & \green{-7.27} & \textbf{211} \\ 
 & pravokotna \ding{55} & \red{1.14} & 451(2.6x) & \red{1.14} & 994(5.6x) & \red{1.39} & 13764(78.2x) & \red{2.97} & 649(3.7x) & \red{1.14} & 404(2.3x) & \green{1.78} & 314(1.8x) & \red{0.75} & \textbf{176} \\ 
\bottomrule
\multicolumn{16}{p{0.95\linewidth}}{\scriptsize \textbf{(naša) -- naše predlagane mere}, B -- Bhattacharyjeva razdalja, JS -- Jensen-Shannonova divergenca, W -- Wassersteinova razdalja, eGauss+ -- razmerje volumnov eGauss+~\cite{Skrjanc_2020}, aB -- Bhattacharyjeva razdalja + aritmetično povprečje + lndet(), O -- Mera prekrivanja, $\tilde{\mathrm{O}}$ -- Mera prekrivanja + Hadamardova neenakost} \\
\end{tabular}
\caption{Primerjava mer prekrivanja rojev v različnih scenarijih prekrivanja za nizke in visoke dimenzije. Vrednosti mer so označene z zeleno, kadar je detekcija pravilna. Računski časi so podani v mikrosekundah ($\mu$s) in kot večkratnik najboljšega rezultata.}
\label{tab_metrics}
\end{table*}
\par V drugem delu smo preverili združevanje rojev z našo mero, pri čemer smo obravnavali prekrivanje večjega števila medsebojno prekrivajočih se rojev. Ti so bili ustvarjeni z naključnim središčem in kovariančno matriko. Kandidati za združitev so bili izbrani s pomočjo mere divergence, njihovi indeksi pa so bili vključeni v matriko povezav, ki se je nato uporabila za iskanje največjih skupin. Pri tem je bil vsak roj prisoten natanko v eni skupini. Vse skupine rojev so se nato združile v enem koraku. Naključno smo generirali roje in vizualno opazovali, ali so bile ustrezno izbrane in združene skupine medsebojno prekrivajočih se rojev.
\section{Rezultati in diskusija}
\par V prvem eksperimentu smo analizirali različne mere prekrivanja v različnih scenarijih. Rezultati eksperimenta so zbrani v tabeli \ref{tab_metrics}. Vse mere so uspešno zaznale prekrivanje in ne-prekrivanje rojev, kar je bilo pričakovano, saj so te mere pogosto uporabljene v ta namen. Ključna razlika med njimi je, da nekatere mere normalizirajo razdalje glede na velikost združenega roja in upoštevajo razmerje med roji, neodvisno od velikosti primerjanih rojev. Pri analizi majhnih rojev je pomembno, da vrednost mere jasno razlikuje med primeri prekrivanja in ne-prekrivanja, saj pričakujemo, da mera jasno pokaže prekrivanje. Ta zahteva je izpolnjena pri vseh naših merah ter pri meri eGauss+. Zanimivo je, da naša mera, ki temelji na razdalji Bhattacharyya, vrne zelo podobne vrednosti kot ta razdalja, s to razliko, da je sposobna detektirati majhne roje.
\par V zadnjem scenariju smo obravnavali dva roja z ortogonalnimi dominantnimi lastnimi vektorji. Združitev teh rojev rezultira v novem roju, ki je bistveno večji od obeh izhodiščnih. V takem primeru združevanje rojev ni zaželeno. Ta scenarij uspešno zaznavata mera eGauss+ in naša mera prekrivanja, ki ne vključuje ocene zgornje meje determinante.
\par Predlagana mera zazna vse scenarije tako pri nizkih kot pri visokih dimenzijah rojev in je pri tem hitrejša od primerjanih metod. Razdalja Wassersteina vključuje koren matrike, kar je računsko zahtevna operacija. Poleg tega ni normirana glede na definicijsko območje značilk, kar predstavlja veliko težavo pri izbiri praga za združevanje. Zanimivo je, da so vse vrednosti divergence Jensen-Shannon in razdalje Bhattacharyya enake za vse primere, kar je nepričakovan rezultat. Verjetno je to posledica izbire povprečne vrednosti za središče in kovariančno matriko, kar je običajen pristop. To potrdi tudi analitična izpeljava:
\begin{align}
    D_{JS}&(P {\parallel} Q) = \notag \\ &
    \frac{1}{4}\Bigg(  
\bigg[\operatorname{tr}\big(\tensor{\Sigma}_M^{-1}  \tensor{\Sigma}_P\big)  + \operatorname{tr}\big(\tensor{\Sigma}_M^{-1}  \tensor{\Sigma}_Q\big) - 2\D\bigg] + \notag\\ &  
    \big[(\tensor{\mu}_M - \tensor{\mu}_P)^\top \tensor{\Sigma}_M^{-1} (\tensor{\mu}_M - \tensor{\mu}_P) + \notag \\&  
    (\tensor{\mu}_M - \tensor{\mu}_Q)^\top \tensor{\Sigma}_M^{-1} (\tensor{\mu}_M - \tensor{\mu}_Q)\big] + \notag \\& 
    \bigg[\ln \Big(\frac{\operatorname{det} \tensor{\Sigma}_M}{\operatorname{det} \tensor{\Sigma}_P}\Big) + \ln \Big(\frac{\operatorname{det} \tensor{\Sigma}_M}{\operatorname{det} \tensor{\Sigma}_Q}\Big)\bigg] \Bigg) = \notag  \\ & = 
    \frac{1}{4}\Bigg(    
\bigg[\operatorname{tr}\Big(\tensor{\Sigma}_M^{-1} \frac{2(\tensor{\Sigma}_P {+} \tensor{\Sigma}_Q)}{2}\Big) - 2\D\bigg] + \notag \\ & 
    2\Big(\frac{\tensor{\mu}_P - \tensor{\mu}_Q}{2}\Big)^\top \tensor{\Sigma}_M^{-1} \Big(\frac{\tensor{\mu}_P - \tensor{\mu}_Q}{2}\Big) + \notag \\ & 
    \bigg[2\ln\operatorname{det} \tensor{\Sigma}_M
    {-} \frac{1}{2}\big(\ln\operatorname{det} \tensor{\Sigma}_P + \ln\operatorname{det} \tensor{\Sigma}_Q\big)\bigg]\Bigg) =
    \notag  \\& = 
    0 + \frac{1}{8}(\tensor{\mu}_P - \tensor{\mu}_Q)^\top \tensor{\Sigma}_M^{-1} (\tensor{\mu}_P - \tensor{\mu}_Q) + \notag \\& \quad \frac{1}{2}\ln\Bigg(\frac{\operatorname{det} \tensor{\Sigma}_M}{\sqrt{\operatorname{det} \tensor{\Sigma}_P \operatorname{det} \tensor{\Sigma}_Q}}\Bigg) = D_B(P{\parallel}Q).
\end{align}
\par
\begin{figure}[!htb]
    \begin{center}
        \includegraphics[width=1\linewidth]{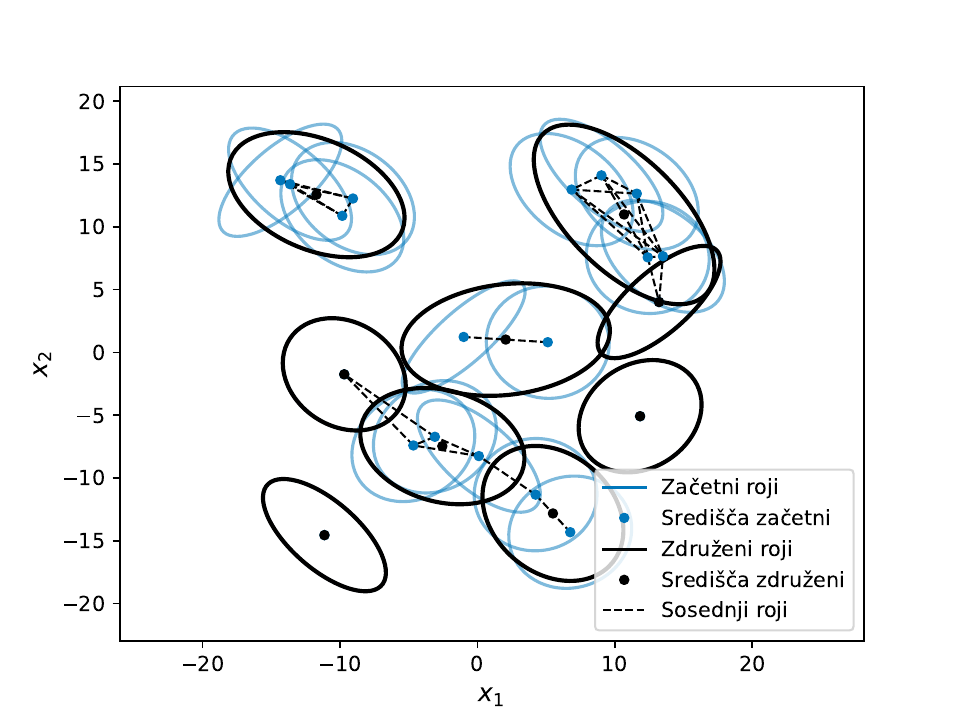}
        \caption{Združevanje več naključno ustvarjenih rojev, kjer smo za izbiro prekrivajočih se rojev uporabili našo mero prekrivanja in roje združili s predlagano enačbo za združevanje več rojev naenkrat.}
        \label{slika_mnozicno_zdruzevanje}
    \end{center}
\end{figure}

 Primeri združenih rojev, dobljeni z enačbo (\ref{eq_zdruzevanje}), so prikazani na sliki \ref{slika_eskperimenti}. Prikazani so ločeni prispevki zaradi kovariančnih matrik in razdalj središč pri skupni kovariančni matriki, kar omogoča dodaten vpogled v delovanje mehanizma združevanja rojev. V večini primerov je razdalja med središči odločilna pri oceni, ali so roji prekrivajoči ali ne, razen v primeru rojev s skupnim središčem. V drugem eksperimentu obravnavamo, ali je mera prekrivanja rojev ustrezna. Kvaliteto mere prekrivanja je mogoče oceniti vizualno na sliki \ref{slika_mnozicno_zdruzevanje}. Vidimo, da predlagana mera ustrezno zazna vse sosednje roje, izbrane so največje skupine medsebojno prekrivajočih se rojev in enačba združevanja več rojev \ref{eq_zdruzevanje} je eksaktna.

\par Kljub temu, da je normalna distribucija najpogosteje uporabljena struktura za opis distribucije verjetnosti, je izračun mer računsko zahteven. Vse mere zahtevajo vsaj izračun determinante, lastnih vrednosti, ali inverza matrike, ki posredno vključujejo izračun determinante. Zanimivo je, da je sam postopek združevanja rojev računsko zelo učinkovit, medtem ko je izračun mere podobnosti, ki določa smiselnost združevanja, precej zahtevnejši. Izračunu ustreznosti združevanja s pomočjo mere prekrivanja se ne moremo izogniti, če imamo na voljo le roje. Predlagani pristop še ne rešuje dokončno problema računske zahtevnosti izračuna prekrivanja rojev, saj vključuje izračun determinante matrik. Kljub temu je predlagani postopek v večini primerov bistveno hitrejši od primerljivih metod.
\section{Zaključek}
\par V tem članku smo predstavili novo mero prekrivanja Gaussovih rojev in predlagali nov postopek izbire in združevanja večjega števila rojev. Motivacija za to raziskavo izhaja iz sprotnega nenadzorovanega učenja, vendar je mogoče predstavljene algoritme uporabiti za združevanje rojev, ki jih dobimo s poljubno metodo rojenja, ki uporablja Gaussove roje kot prototipe. Predlagana mera je večkrat hitrejša od drugih metod prekrivanja in je sposobna pravilno zaznati prekrivajoče in neprekrivajoče roje v primerih, kjer druge metode odpovejo. V prihodnosti bodo ti pristopi uporabljeni za združevanje modelov v sodelovalnem učenju (Federated Learning), kjer vsak lastnik podatkov zgradi svoj lokalni model rojev, ti pa se združijo v globalni model brez posredovanja podatkov. Primer aplikacije takega sistema je sprotno rojenje podatkov goljufivih transakcij v finančnih sistemih ali zaznavanje vdorov v spletni varnosti.
\small
\bibliographystyle{IEEEtran}
\input{erk.bbl}

\end{document}